# Efficient Toxicity Prediction via Simple Features Using Shallow Neural Networks and Decision Trees

Abdul Karim,*,†,§ Avinash Mishra,*,†,‡,§ M. A. Hakim Newton,*,† and Abdul Sattar*,†

†Institute of Integrated and Intelligent Systems, Griffith University, Nathan 4111, Australia
‡Novo Informatics Private Limited, New Delhi 110049, India

  Supporting Information

**ABSTRACT:** Toxicity prediction of chemical compounds is a grand challenge. Lately, it achieved significant progress in accuracy but using a huge set of features, implementing a complex blackbox technique such as a deep neural network, and exploiting enormous computational resources. In this paper, we strongly argue for the models and methods that are simple in machine learning characteristics, efficient in computing resource usage, and powerful to achieve very high accuracy levels. To demonstrate this, we develop a single task-based chemical toxicity prediction framework using only 2D features that are less compute intensive. We effectively use a decision tree to obtain an optimum number of features from a collection of thousands of them. We use a shallow neural network and jointly optimize it with decision tree taking both network parameters and input features into account. Our model needs only a minute on a single CPU for its training while existing methods using deep neural networks need about 10 min on NVidia Tesla K40 GPU. However, we obtain similar or better performance on several toxicity benchmark tasks. We also develop a cumulative feature ranking method which enables us to identify features that can help chemists perform prescreening of toxic compounds effectively.

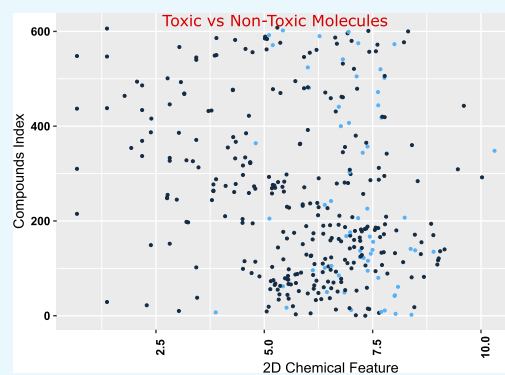

## ■ INTRODUCTION

Chemical toxicity is an important measure in environmental, agricultural, and pharmaceutical science.[1,2] In the environmental context, toxic chemicals may cause varieties of chronic diseases.[3] In pharmacology, toxicity prediction plays a vital role in the drug discovery pipeline.[4,5] This makes toxicological screening to be mandatory for the development of new drugs and for the extension of the therapeutic potential of existing molecules.[6] Several in vitro/in vivo techniques have been devised to determine varieties of toxic effect including eye irritancy test,[7,8] mutagenicity,[9,10] toxicokinetics,[11] neurotoxicity,[12] embryotoxicity,[13] and genetic toxicity.[14] However, these techniques for examining chemical toxicity are highly cost- and time-intensive.[15] In addition, ethical concerns have also been raised against the usage of animals for toxicology screening.[16] Therefore, there is an increased demand for cost- and time-efficient toxicological screening methods.[17] In silico approach for toxicity prediction has clearly opened a new avenue to address the initial round of screening with acceptable accuracy.[18] These in silico methods are primarily modeled using high throughput screening assays on various tasks of toxicity.[19]

In recent years, machine learning methods have been widely used in drug discovery.[20] Under the umbrella of machine learning, methods such as k-nearest neighbors (KNN) and support vectors machines (SVM) were used for structure activity relation (SAR) techniques.[21−23] Performance of traditional machine learning algorithms depends heavily upon the quantity and quality of training data along with domain knowledge-based feature engineering. For instance, a KNN model used for hazard evaluation support systems was designed on carefully selected eight fingerprints as input features for a relatively small data set of 94 chemicals in the training set and 24 chemicals in the test set.[24] Similarly in another study, 74 topological descriptors with 314 training instances were used for specific COX-2 inhibitors.[22] These models perform relatively better on smaller data sets with fewer preselected features. One key limitation of KNN algorithm is the exponential rise of computational cost with the size of the input samples.[25−27] In contrast, nonlinear SVMs can manage high dimensional data but do not exhibit sufficiently robust performance on diverse chemical descriptors.[28]

Besides KNN and SVMs, naive Bayes and random forest (RF) methods were also used extensively for toxicity prediction.[29−32] Although RF is a decision tree (DT) method capable of handling high-dimensional and diverse features, yet in many cheminformatics data sets, it shows a relatively low classification accuracy when compared to deep neural networks (DNNs).[28,33] DNN is an artificial neural network with more than one hidden layer between the input and output while a shallow neural network (SNN) has only one hidden layer.[34−37] In order to achieve high accuracy in a DNN, relatively a large data set is preferred with numerous features.[38,39] In RF, features are used in raw form while DNN converts them to complex features using hidden layers.[33,40] Moreover, hyperparameter tuning in DNN gives a better control over a granular level optimization unlike in other machine learning approaches.

Table 1. NR, SR, and AM Data Division: Train, CV, and Test Sets

| task | train | toxic/non-toxic | CV | toxic/non-toxic | test | toxic/non-toxic |
|---|---|---|---|---|---|---|
| NR-AHR | 7863 | 937/6926 | 268 | 30/238 | 594 | 73/521 |
| NR-AR | 9036 | 374/7950 | 288 | 3/285 | 573 | 12/559 |
| NR-AR-LBD | 8234 | 284/7950 | 249 | 4/245 | 567 | 8/559 |
| NR-aromatase | 6959 | 352/6607 | 211 | 18/193 | 515 | 37/478 |
| NR-ER | 7421 | 916/6505 | 261 | 27/234 | 505 | 50/455 |
| NR-ER-LBD | 8431 | 415/8016 | 283 | 10/273 | 585 | 20/565 |
| NR-PPARG | 7883 | 193/7690 | 263 | 15/248 | 590 | 30/560 |
| SR-ARE | 6915 | 1040/5875 | 230 | 47/183 | 540 | 90/450 |
| SR-HSE | 7879 | 386/7493 | 263 | 10/253 | 594 | 19/575 |
| SR-MMP | 7071 | 1117/5954 | 234 | 38/196 | 530 | 58/472 |
| SR-p53 | 8349 | 509/7840 | 265 | 28/237 | 601 | 40/561 |
| SR-ATAD5 | 8775 | 317/8458 | 268 | 25/243 | 606 | 36/570 |
| AM | 3900 | 2097/1803 | 1300 | 699/601 | 1300 | 699/601 |

DNN attracted considerable attention in chemical information modeling community when Ma et al. won the "Merck Molecular Activity Challenge" using DNN networks in predicting the biomolecular target for a drug.[41,42] Later in 2014, "Tox21 Challenge" was also won by a group who used DNN.[33] Following this trend, many other groups in computational chemistry used DNN models to achieve high accuracy to predict various chemical and biological characteristics including toxicity,[43,44] activity,[45−47] reactivity,[48−50] solubility,[51] ADMET,[52] docking,[53] and QM-compound energies.[44,54,55] Even after achieving the state-of-the-art accuracy in various cheminformatics tasks, limited model interpretability of DNN made it less preferred in real world health informatics applications.

An ideal model is characterized by its high accuracy, capability to deal with molecular descriptor diversity, ease of training, and slightly more importantly interpretability.[56] Unfortunately, most machine learning approaches act like "black box"; which means no insights are available from them about the problem or the solution structures, making them less trustworthy from human perspective. In this paper, we strongly argue for the models and methods that are simple in machine learning characteristics, efficient in computing resource usage, and powerful to achieve very high accuracy levels. We therefore present a novel hybrid framework that uses DT and SNN to build a simple machine learning model that paves a path to feature interpretability while achieving similar or better accuracy by selecting only the relevant features to train the model.

Using the proposed hybrid framework, we then construct a single task (ST) prediction model and train it on nuclear receptor (NR), stress response (SR), ames mutagenicity (AM), *Tetrahymena pyriformis* IGC50, oral rat LD50, 96 h fathead minnow LC50 data set (LC50 set), 48 h Daphnia magna LC50 data set (LC50-DM set), toxicity data sets. For all the toxicity data sets considered in this study, we calculate only 2D chemical descriptors, which are less multifarious in nature and easy to calculate. The SNN in our model has only one hidden layer with 10 neurons and is trained with significantly fewer features (in the range of hundreds) than existing methods. The training time for our prediction models is reduced to ≈1 min on Intel Core i5 CPU, whereas the same was reported ≈10 min in the previous study using NVidia Tesla K40 GPU.[57] However, our model still achieved relatively better or competitive average accuracy for SR, NR, AM, IGC50, and LD50 moderate size data sets.

It is worth noting that our main objective is not merely to improve the accuracy, but also to focus more on the compute intensiveness, obtaining simpler prediction models in terms of numbers of features used and architecture of the neural network and stepping toward interpreting the decisions made by the model. As a proof of the concept, we developed a cumulative feature ranking method to elucidate the interpretation of the descriptors that are the most responsible for NR, SR, and AM toxicity types. Similar analysis can be extended to other types of toxicity data sets. These descriptors showed high classification strength to discriminate toxic compounds and could be used as initial indicators for detecting NR, SR, and AM toxicity types.

■ RESULTS

In this section, we discuss the benchmark data sets and performance on three case studies and final test sets, investigate prediction potential of 2D descriptor, analyze the comparative landscape, and explain feature interpretability of our classification results.

**Benchmark Classification Data Sets.** NR and SR data sets were collected from Tox21 challenge.[58] NR assays were classified into subtasks pathways: (1) aryl hydrocarbon receptor, (2) androgen receptor-full, (3) androgen receptor-luciferase, (4) aromatase, (5) estrogen receptor (ER) alpha, (6) ER alpha-luciferase, and (7) peroxisome proliferator-activated receptor gamma. SR assays were classified into 5 subtasks pathways: (1) antioxidant response element, (2) heat shock response/unfolded protein response, (3) mitochondrial membrane potential (MMP), (4) DNA damage p53 pathway, and (5) genotoxicity indicated by ATAD5. A separate benchmark data set for AM was also obtained.[59] It should be noted that for SR and NR, the data was predivided into training, held out cross validation (CV), and separate test sets by the Tox21 repository. For AM, no such division was given, so we divided it into train (60%), CV (20%), and test (20%) sets randomly. Later, the CV and train were mixed together for $k$-fold CV ($k$ = 5) analysis; however, the test sets were kept held out. Each set contains toxic and nontoxic compounds, and the detailed description is provided in Table 1. It should be noted that Table 1 mentions the data setting after the cleaning and quality control.

**Prediction Potential of 2D Descriptors.** Representation of chemical compounds in 2D form as a connection table is used to calculate their 2D descriptors. These descriptors are relatively easier to calculate and computationally less intensive. PaDEL descriptor tool was used to calculate 1422 2D descriptors.[60] Names and short descriptions of these 2D descriptors are provided in Table S1. Primarily, prediction (classification) potential of these features was evaluated by performing a dry run using a neural network model on training data set of each task (2nd column of Table 1). Training sets for NR and SR were up-sampled and split into internal training/validation set with 70/30 ratio for examining the prediction potential of 2D features. Here, the CV and the test set (4th and 6th column of Table 1) were not considered, as the aim was not to build a final prediction model rather to estimate the prediction power of the 2D features. AUC-ROC for each toxicity task of NR and SR was calculated using internal validation set as shown in Figure 1. It showed clearly in Figure 1 that 2D features have high potential to discriminate toxic and nontoxic compounds.

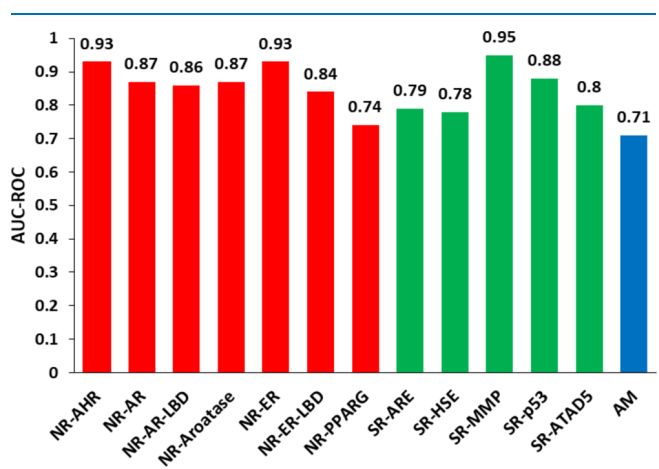

**Figure 1.** Area under the curve (AUC-ROC) for three toxicity data sets, calculated on the internal test set to evaluate the prediction potential of 2D features.

The highest AUC-ROC of 0.95 was obtained for MMP task which belongs to SR panel while ER from NR panel showed the lowest AUC-ROC of 0.74. Although AUC-ROCs shown in Figure 1 is overestimated as it is on the internal validation set created from training set of Tox21, but it clearly shows a good performance of 2D descriptors as features in the prediction model. Thus, the results shown in Figure 1 confirmed that 2D descriptors alone have the potential to discriminate between toxic and nontoxic compounds for NR and SR signaling pathways. The same procedure was repeated for AM data set as well and the AUC-ROC is included in the figure. It should be noted that results shown in Figure 1 are not the final results, instead it shows that there is a prediction potential in 2D features for all the three toxicity tasks. The results shown in Figure 1 were obtained without any feature optimization and no hybrid framework of neural network and DT is used. This result could be improved with proper optimization as discussed in latter sections.

**Case Study-I: Series Versus Parallel Optimization.** Our hybrid model is composed of two main components, that is a SNN and a DT classifier (detail is given in Methods section). Optimization of different parameters involved in two components of our hybrid framework is an essential phase to achieve a high accuracy. Parameters of both components (i.e., DT and SNN) of the hybrid model could be optimized simultaneously (parallel mode) or one after another (series mode). A case study was conducted to compare the performance of series and parallel optimization on SR, NR, and AM data sets. ER task of NR class has shown the lowest accuracy in earlier studies by different groups under Tox21 challenge while MMP of SR class has showed the best result.[33] Thus, NR-ER, SR-MMP, and AM were selected for this case study. Similarly, two most critical parameters, one from the DT and the other one from the neural network, were selected for optimization.

Threshold is an important parameter of a DT classifier that sets cut-off value for the selection of features and "dropout" refers to dropping out units in hidden layers of the neural network to prevent overfitting. These two parameters one from DT and one from neural network were optimized in series and parallel mode using the grid search technique considering AUC-ROC for the randomly chosen 20% cross-validation set as an objective function. The process of obtaining the cross-validation set is explained in the latter sections. It should be noted that the same process is repeated to obtain the cross-validation set for upcoming case studies as well. In addition to "threshold" in the DT classifier, the number of trees (n_estimator)[61,62] was also tested in the range of 10–2000 on selected tasks, that is NR-ER, SR-MMP, and AM. Figure 2a shows the behavior of n_estimator with the AUC-ROC on cross-validation set. Initially AUC-ROC showed ripples but then it became stable after 1000 number of trees. This suggests that n_estimator could be fixed to 1000 to make the model robust. Once the number of trees was fixed, threshold values were taken in grid search over [0.08, 0.09, 0.1, 0.2, 0.3, 0.4, 0.5, 0.6, 0.7, 0.8, 0.9, 1.0, 1.1, 1.2, 1.3, 1.4, 1.5, 1.6, 1.7, 1.8, 1.9, 2.0, 2.1, 2.2, 2.3] range while dropout was taken over [0, 0.1, 0.2, 0.3, 0.4, 0.5, 0.6, 0.7, 0.8, 0.9]. During series mode, threshold was optimized first for the AUC-ROC of CV set which resulted in the optimized value of 1.6, 1.1, and 1.5 for NR-ER, NR-MMP, and AM, respectively. Later, with these optimized threshold values selected, the dropout parameter was optimized in its search space. Optimum values for the dropout were 0.4, 0.2, and 0.1 with AUC-ROC 0.811, 0.949, and 0.864 for NR-ER, SR-MMP, and AM, respectively. Hence, these values were considered as optimized values for threshold and dropout. In parallel mode optimization, each combination of threshold and dropout were explored simultaneously and respective AUC-ROCs were calculated.

The parallel optimization resulted in several pairs of values (1.6, 0.7), (1.2, 0.4), and (1.3, 0.3) for threshold and dropout with the best AUC-ROC of 0.789, 0.946, and 0.846 for NR-ER, SR-MMP, and AM respectively. Results of series and parallel optimization are shown in Figure 2b. In all the three cases, series and parallel optimizations perform very close to one another based on their AUC-ROC. However, the series mode achieved marginally higher AUC-ROC than the parallel mode. Additionally, the parallel optimization between two or more parameters from DT and SNN was found to be compute-intensive. This concluded to deployment of series parameter optimization across the components of hybrid framework (DT and SNN) to build our predication model.

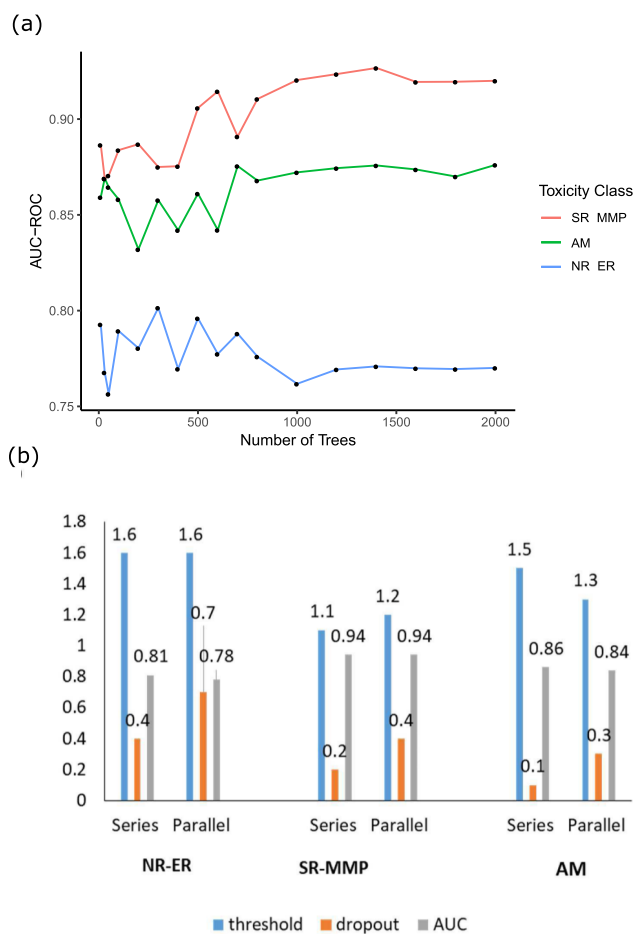

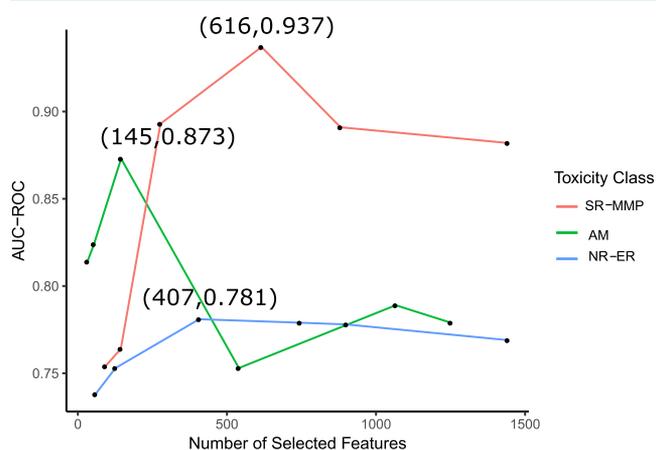

**Figure 3.** Selected number of effective 2D features used to build prediction models to achieve higher AUC-ROC for NR-ER, SR-MMP, and AM toxicity classes.

**Figure 2.** Variable parameters of hybrid learning model for series and parallel optimization to achieve betters results. (a) Number of trees in DT classifier vs AUC-ROC for NR-ER, SR-MMP, and AM. (b) Selected values of threshold and dropout are shown in series and parallel optimization for NR-ER and SR-MMP task with AUC-ROC as an objective function.

**Case Study-II: Do We Really Need a Large Set of 2D Features?** In this case study, we wanted to know the number of 2D features which are sufficient for very good performance. This case study was inspired by a theorem called "curse of dimensionality", which states that beyond a certain point, the inclusion of additional features may lead to higher probabilities of error.[39] Moreover, there is a need of reducing the number of features to make the model simple and less compute intensive. The reduced number of features should be nearly optimum for a good performance and thus may help in feature interpretability. It should be noted that in this work, the term "nearly optimum" is referring to the reduced number of 2D features which can give better performance while searching over different values of thresholds in DT component of our hybrid model.

We plotted the AUC-ROC of all the three toxicity data sets (for cross-validation sets) against the number of features selected to know whether we can achieve better performance with fewer number of features. The threshold value of a DT classifier component was varied over a space of [0.0, 0.5, 1.0, 1.5, 2.0, 2.5]. The greater the threshold value, the lesser the number of features selected. The details of how the threshold changes the number of features selected are given in the Methods section. A SNN was trained for different numbers of selected features and the results are shown in Figure 3. In each case, we see that for better performance, we need not train our model using all available 2D features, but instead a reduced number of features are sufficient to get the better performance. In case of AM in Figure 3, only 145 selected features achieved the highest AUC-ROC on cross-validation set. If we further increase the number of features, the performance degrades. A similar trend can be seen for NR-ER and SR-MMP as well, although the performance does not degrade much with the increase of number of features. In these cases, it is a better choice to select the smaller number of features to make the model simple, less compute-intensive and improve feature interpretability.

Considering case studies I and II, we developed a hybrid model (explained in the Methods section) which enables the SNN to select the small number of effective features (nearly optimum) to be trained on while jointly optimizing parameters of the SNN and a DT.

**Case Study-III: Why a SNN?** The use of SNN as one of the components in our hybrid model was motivated by "universal approximation theorem". It states that a SNN (having one hidden layer with finite number of neurons) is a universal function approximator.[63] However, in practice a DNN performs better on a large set of raw features. As the feature selection module of our hybrid model effectively selects a reduced number of features, we expect that a simple SNN with a small number of neurons will be able to perform better or similar to a DNN. The main idea was to make a hybrid model in which a SNN would extract relevant knowledge (effectively selecting the features to be trained on) from the DT classifier. In order to know if a SNN will perform similar or better on selected features, we performed another case study in which the number of hidden layers was varied from 1 to 5 for all three types of toxicity data sets as shown in Figure 4. It was found that a neural network with one hidden layer has relatively higher AUC-ROC on cross-validation set (on selected features) than a DNN for all three toxicity data sets. This concludes the implementation of a SNN for better results.

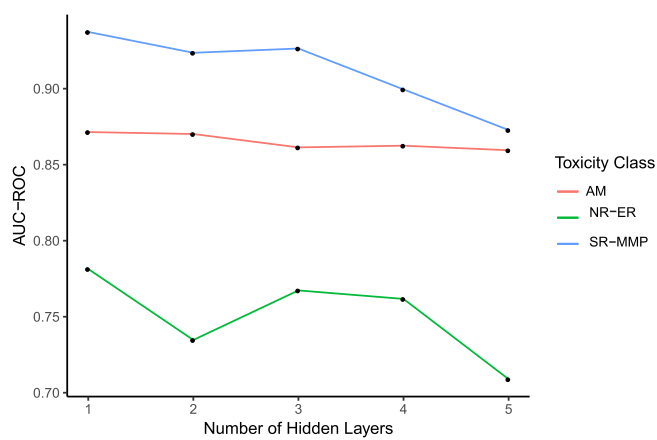

**Figure 4.** Classification performance with varying number of hidden layers used in neural network, for NR-ER, SR-MMP, and AM toxicity classes.

**Final Test Sets and Cross-Validation Performance.** The hybrid model was evaluated using the final test sets for each of the three toxicity data sets (7 tasks for NR, 5 for SR, and 1 for AM) shown in Table 1 (6th column). The test set was not part of the training or validation, and hence it was considered as blind testing of the prediction model. Each toxicity task has specific parameter values as shown in Table S2. These parameters for the individual tasks were optimized using 5-fold CV. The Table 1 (2nd and 4th column) were combined together and then divided into 5 sets; 4 of which were used to train the model and 5th to validate the model. Fragment similarity-based sampling method was used to divide the data to make sure that active and non-active compounds ratio is constant in all 5 sets. AUC-ROC (area under the receiving operating curve), AUC-PR (area under the precision−recall curve) also called average precision, and F1 score were used to evaluate the performance of the classifier. F1 score is computed over a range of thresholds and maximum F1 score is reported. The values for all the performance metrics for all classification tasks on the final tests and cross-validation sets are given in Table 2. It also shows the number of features selected which is a subset of the total ≈1422 features for NR and SR while ≈1249 for AM to build each model. In order to evaluate the model on the basis of true positive rate (TPR), false positive rate (FPR), and precision−recall, we provide plots of receiving operating curves and precision recall curves for cross-validation in Figure 5 and final test sets in Figure 6.

In order to reduce the chance of error in the final result and to show robustness of our hybrid model, we performed ensemble averaging on all three toxicity data sets. The result of each ensemble model (total of 4 for each task within individual data set) is given in Table S2. We also developed RF and SVM based models for all three classification toxicity tasks and report the AUC-ROC on the final test data using 2D features given in Table S2.

**Comparative Landscape on the Tox21 Bench-Mark External Test.** On SR and NR data, our hybrid model was compared with the winning model of Tox21 challenge.[33,57] We outperformed other methods in AUC-ROC for SR and NR toxicity data sets. The winning model of Tox21 challenge is based on DNN and is trained on ≈273 577 features for NR and SR data sets using a multitask approach. In this approach, DNNs up to four layers with thousands of neurons in each layer were tested. By harnessing the ability of a DNN to create intermediate complex features for prediction, they were able to achieve the average AUC-ROC of 0.826 for NR and 0.858 for SR on the final test set. Training of the model was computationally very expensive and took ≈10 min to train on NVIDIA Tesla K40 GPU. The large numbers of features used by the model made it very hard to interpret which features are playing a vital role in decision making.[57]

The second ranked team, AMAZIZ, developed consensus models using associative neural network (ASNN) to achieve an average AUC-ROC of 0.816 for NR and 0.854 for SR. ASNN represents a combination of an ensemble of feed-forward neural networks and the KNN technique.[64] The information about the total number of features used and the training time is not reported.[64] The third ranked group, dmlab, developed ensemble models with combining various fingerprinting tools using RF and extra tree (ET) classifier to achieve an average AUC-ROC of 0.811 for NR and 0.850 for SR.[65] After Tox21 challenge, other groups developed prediction models for NR and SR data sets.[66−68] Chemception developed convolutional neural networks to predict toxicity using 2D images of compounds without explicitly calculating chemical descriptors and achieved average AUC-ROC 0.787 for NR and 0.739 for SR.[68] Capuzzi et al. used DNN with an ensemble of 2489 molecular descriptors to achieve a very good overall average AUC-ROC of 0.840 for both NR and SR.[66] SMILES2vec used deep recurrent neural networks that automatically learn features from the SMILES data, and the reported average AUC-ROC is 0.799 for both NR and SR.[67]

**Table 2. Performance on Final Test Sets and Cross-Validation Set (5-Fold) for NR, SR, and AM Toxicity**[a]

| task | features selected | final test AUC-ROC | final test AUC-PR | final test max F1 score | cross-valid AUC-ROC | cross-valid AUC-PR | cross-valid max F1 score |
|---|---|---|---|---|---|---|---|
| NR-AHR | 270 | 0.921 | 0.642 | 0.633 | 0.911 | 0.611 | 0.622 |
| NR-AR | 284 | 0.743 | 0.273 | 0.563 | 0.842 | 0.620 | 0.655 |
| NR-AR-LBD | 365 | 0.881 | 0.155 | 0.210 | 0.900 | 0.726 | 0.754 |
| NR-aromatase | 815 | 0.794 | 0.324 | 0.357 | 0.898 | 0.431 | 0.513 |
| NR-ER | 292 | 0.822 | 0.471 | 0.535 | 0.781 | 0.524 | 0.516 |
| NR-ER-LBD | 755 | 0.836 | 0.375 | 0.410 | 0.876 | 0.657 | 0.666 |
| NR-PPARG | 528 | 0.858 | 0.310 | 0.377 | 0.859 | 0.500 | 0.583 |
| NR average | 472 | 0.836 | 0.364 | 0.440 | 0.867 | 0.581 | 0.615 |
| SR-ARE | 615 | 0.828 | 0.493 | 0.563 | 0.850 | 0.520 | 0.603 |
| SR-HSE | 1028 | 0.832 | 0.494 | 0.565 | 0.892 | 0.495 | 0.619 |
| SR-MMP | 685 | 0.958 | 0.700 | 0.810 | 0.943 | 0.813 | 0.740 |
| SR-p53 | 223 | 0.875 | 0.305 | 0.382 | 0.910 | 0.640 | 0.647 |
| SR-ATAD5 | 390 | 0.820 | 0.301 | 0.370 | 0.832 | 0.476 | 0.530 |
| SR average | 588 | 0.862 | 0.458 | 0.538 | 0.885 | 0.588 | 0.629 |
| AM | 145 | 0.878 | 0.910 | 0.827 | 0.879 | 0.890 | 0.821 |

[a]AUC-ROC is the area under the ROC curve and AUC-PR is the area under the curve of precision−recall curve.

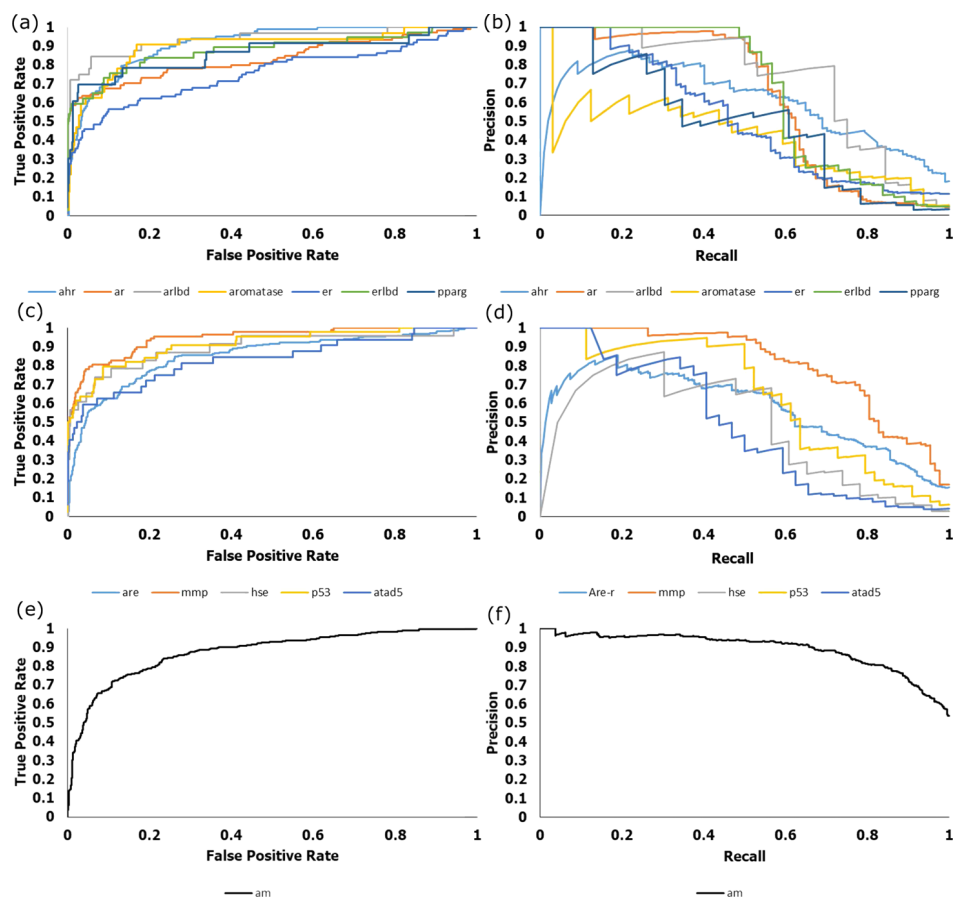

**Figure 5.** TPR and FPR plot for cross-validation sets (a,c,e). Precision−recall plot for cross-validation sets (b,d,f).

Our hybrid framework used reduced number of simple (easy to compute) 2D features to achieve the state of the art average AUC-ROCs. In contrast to other methods, we used SNN (1 hidden layer, 10 neurons) that makes the model computationally efficient and opens the avenue for interpretability. The average training time for our hybrid framework method is always less than a minute for all the tasks. Effectively reduced number of features selected in an optimization loop using DT and SNN improves the model to achieve the highest accuracy. Table 3 shows the comprehensive comparison of our model with others for SR and NR. In addition to accuracy, we also compared our method on model complexity ground with top 5 models in Tox21 challenge. Table 4 shows methods, training time, and number of feature for top 5 models of Tox21 challenge and for AM benchmark data set. DeepTox model achieved AUC-ROC close to our method but it used DNN with 273 577 features. Tables 3 and 4 jointly demonstrate the performance of our hybrid framework on accuracy and complexity verticals.

Cytotoxicity is the major concern for determining true chemical toxicity of any compound, and it adds noise to data in terms of false positive. Tox21 data were examined experimentally by Tox21 group for cytotoxicity effects. Here, eight tasks were tested on various cell lines for cytotoxicity check. These tasks are (1) AHR, (2) AR, (3) ARE, (4) aromatase, (5) ER, (6) HSE, (7) p53, and (8) PPARG. Bla and mda cell lines were used for AR, while bla and bg1 cell lines were used for ER to determine agonist and antagonist effect. However, other tasks were tested on single cell line, they are AHR → HepG2, ARE → bla, aromatase → MCF7 ere, HSE → bla, p53 → bla, and PPARG → bla. These information data on cytotoxicity assays are provided at https://tripod.nih.gov/tox21/. On each cell line, a given compound is tested multiple times in the range of 6−204 replicates to calculate AC50. We found that compound behaves similar in their every run. This implies that if a compound is cytotoxic then it is always detected as cytotoxic throughout their multiple experimental runs, within the same cell line as well as across the different cell lines. However, AHR, ARE, HSE, and p53 did not show any such cytotoxic compound, so these are free from cytotoxicity. On the other hand, AR, ER, aromatase, and PPARG tasks have cytotoxic compounds tested in different cell lines. Further, we determined the significance of number of cytotoxic compounds with respect to their complete dataset. We performed statistical $t$-test to evaluate the cytotoxicity significance. Two populations, (i) with cytotoxic compounds and (ii) without cytotoxic compounds were supplied with their AC50 values to $t$-test at 95% confidence interval ($\alpha = 0.05$). Two AR tasks in different conditions (mda cells agonist and mda cells antagonist) and one ER (bg1 agonist) showed nonsignificance effect of number of cytotoxic compounds with respect to complete dataset, $p$-values recorded are 0.9568, 1 and 0.9155, respectively. Other than these three, all others tasks in different cell lines showed a significant number of cytotoxic compounds in their dataset (i.e. $p$-value < 0.05). None of the toxicity prediction method on Tox21 dataset addressed these cytotoxic effects. Therefore, we also did not remove these data points during our AUC-ROC calculation in order to make fair comparison on same size of dataset.

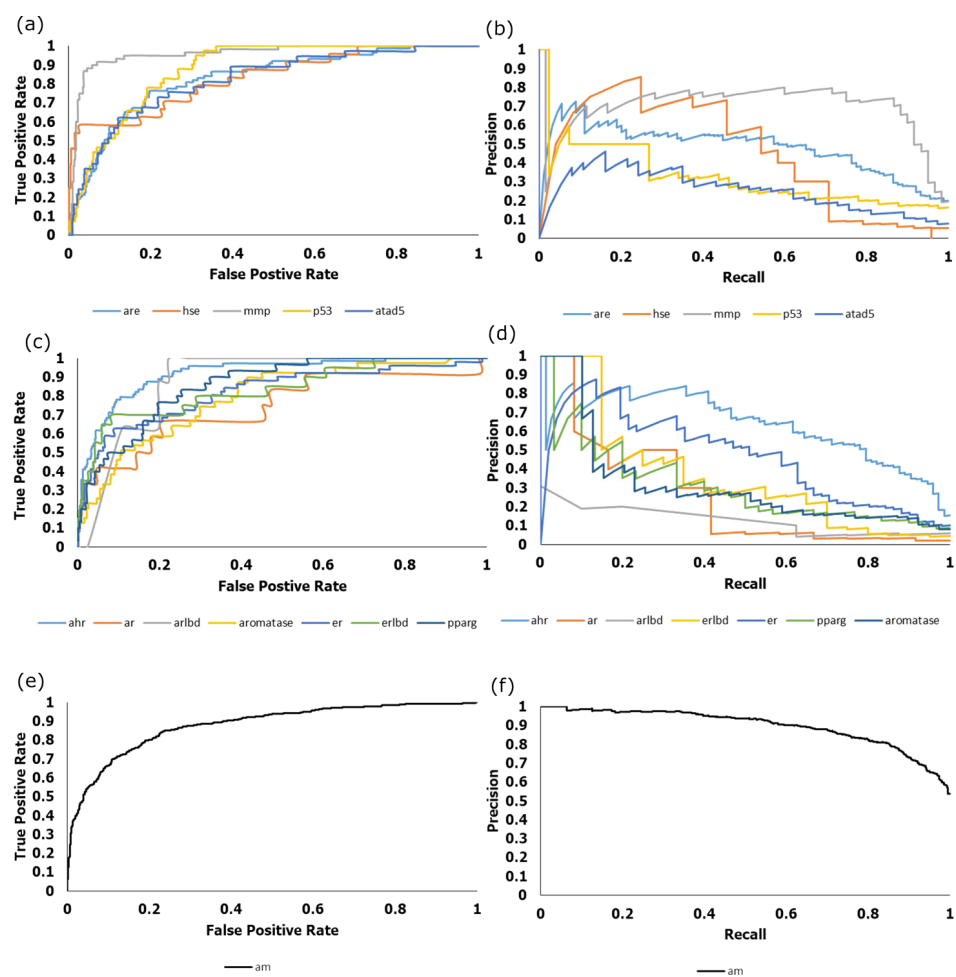

**Figure 6.** TPR vs FPR plot for test sets (a,c,e). Precision with recall plot for test sets (b,d,f).

**Table 3. Comparative Analysis of Different Methods Used for NR and SR on the External Standard Tox21 Bench Mark Test Set**

| name | NR average AUC-ROC | SR average AUC-ROC |
|---|---|---|
| **our method** | **0.836** | **0.862** |
| DeepTox[57] | 0.826 | 0.858 |
| AMAZIZ[64] | 0.816 | 0.854 |
| Capuzzi[66] | 0.831 | 0.848 |
| dmlab[64] | 0.811 | 0.85 |
| T | 0.798 | 0.842 |
| microsomes | 0.785 | 0.814 |
| filipsPL | 0.765 | 0.817 |
| Charite | 0.75 | 0.811 |
| RCC | 0.751 | 0.781 |
| frozenarm | 0.759 | 0.768 |
| ToxFit | 0.753 | 0.756 |
| CGL | 0.72 | 0.791 |
| SuperTox | 0.682 | 0.768 |
| kibutz | 0.731 | 0.731 |
| MML | 0.7 | 0.753 |
| NCI | 0.651 | 0.791 |
| VIF | 0.702 | 0.692 |
| toxic avg | 0.659 | 0.607 |
| Swamidass | 0.596 | 0.593 |
| Chemception[68] | 0.787 | 0.739 |
| ProTox-II[69] | 0.794 | 0.850 |

**Table 4. Training Time and Model Complexity of the Top 5 Models from the Tox21 Challenge Bench Mark Data Set**

| task | name | method | number of features | training time | AUC-ROC (test) |
|---|---|---|---|---|---|
| NR | **our method** | **DT + SNN** | 472 | ≈1 min CPU | **0.836** |
| NR | DeepTox[57] | DNN | 273 577 | ≈10 min GPU | 0.826 |
| NR | AMAZIZ[64] | ASNN | NA | NA | 0.816 |
| NR | Capuzzi[66] | DNN | 2489 | NA | 0.831 |
| NR | dmlab[64] | RF + ET | 681 | ≈13 s CPU | 0.811 |
| SR | **our method** | **DT + SNN** | 588 | ≈1 min CPU | **0.862** |
| SR | DeepTox[57] | DNN | 273 577 | ≈10 min GPU | 0.858 |
| SR | AMAZIZ[64] | ASNN | NA | NA | 0.854 |
| SR | Capuzzi[66] | DNN | 2489 | NA | 0.848 |
| SR | dmlab[64] | RF + ET | 681 | ≈13 s CPU | 0.850 |

However, to show the cytotoxic effect separately, we removed these datapoints from complete dataset of their tasks (AR, ER, aromatase, PPARG). Removal of these compounds did not show any major change in AUC-ROC in contrast to earlier calculation (Table 3), although it dropped to a small extent in two cases (AR and PPARG) while improved in one (ER) and remain unchanged in one case (aromatase). AUC-ROC dropped by 0.033 and 0.030 for AR and PPARG, respectively. In case of aromatase, it remained unaffected, while in ER it improved by 0.008.

**Regression Toxicity Prediction.** In order to verify the general applicability of 2D features predictive power and robustness of our model, we performed additional experiments using four regression-based data sets. These data sets namely, 96 h fathead minnow LC50 data set (LC50 set), 48 h Daphnia magna LC50 data set (LC50-DM set), 40 h *T. pyriformis* IGC50 data set (IGC50 set), and oral rat LD50 data set (LD50 set), were obtained from Wu and Wei while the setting (train test split) was kept the same as given in their recent work on toxicity.[70] In this work, Wu and Wei used various types of approaches to verify the predictive power of element specific topological descriptors, auxiliary molecular descriptors (AUX), and a combination of both for the four types of toxicity data sets. They named their predictive model as TopTox. In our case, similar to the three classification data sets discussed earlier, we calculated 2D features using Padel descriptor for these regression based data as well. The parameters of SNN and DT were jointly optimized (explained Methods section) using 5 fold CV. We compared our ST hybrid model results with ST-DNN, multitask DNN (MT-DNN), and consensus models of TopTox as shown in Table 5. For IGC50 data set, our model achieves an $R^2$ value of 0.810, which is better than the TopTox. For oral rat LD50, our model achieves an $R^2$ of 0.629, which is better than MT and ST deep learning models while competitive to $R^2$ of 0.653 that achieved using consensus models.[70] However, on relatively smaller data sets such as LC50 and LC50-DM, our model struggles to achieve better results than best performing models such as MT-DDN and consensus. However, our ST-based hybrid model outperforms in 3 out of 4 ST method of TopTox.

Table 5. Regression Co-Efficient ($R^2$) Comparison of Our Method with TopTox on Regression Based Test Set[a]

| task | data set size | TopTox | | | our method |
|---|---|---|---|---|---|
| | | ST-DNN | MT-DNN | consensus | ST-hybrid |
| IGC50 | 1792 | 0.708 | 0.770 | 0.802 | **0.810** |
| LC50-DM | 353 | 0.459 | **0.788** | 0.681 | 0.616 |
| LC50 | 823 | 0.692 | 0.771 | **0.789** | 0.678 |
| LD50 | 7413 | 0.614 | 0.626 | **0.653** | 0.629 |

[a]ST: single-task method; MT: multitask method.

For smaller data sets, the feature selection method of our model selects relatively higher number of features which have high correlations among them. This is because that the feature selection method in our model is acting as a coarse granular filter for small data sets. Currently, the better optimization of feature selection module is not in the scope of this paper and can be included in the future work.

**5-Fold Cross-Validation Set Evaluation.** Recently, the AdmetSAR method[71−73] showed performance on four Tox21 tasks (AR, ER, aromatase, and PPARG), while ProTox-II[69] used the complete set of Tox21 classification tasks. Table 6 shows the AUC-ROC comparison of AdmetSAR and Protox-II with our method on Tox21 dataset. We achieved better results (11/12 tasks) in comparison to ProTox-II. However, our method performed competitively to AdmetSAR. Additionally, we also compared regression tasks with AdmetSAR as shown in Table 7 (ProTox-II not used regression dataset). Here, we achieved better $R^2$ than AdmetSAR on all regression tasks. On AM set, we compared our results with the state of the art methods.[59,69,71] Our model achieved better AUC-ROC of 0.879 on the benchmark data set as compared to the AUC-ROC of 0.860 achieved by Hansen et al.[59] However, we lagged behind AdmetSAR and ProTox-II on this dataset, as they showed 0.91 and 0.90 AUC-ROC, respectively.

Table 6. AUC-ROC Performance Comparison of Our Method with AdmetSAR and ProTox-II on Tox21 5-Fold CV Data

| task | ProTox-II | AdmetSAR | our method |
|---|---|---|---|
| NR-AHR | 0.89 | NA | **0.911** |
| NR-AR | 0.84 | **0.886** | 0.842 |
| NR-ARLBD | 0.87 | NA | **0.900** |
| NR-aromatase | 0.86 | 0.886 | **0.898** |
| NR-ER | 0.75 | **0.880** | 0.781 |
| NR-ERLBD | 0.85 | NA | **0.876** |
| NR-PPARG | 0.81 | 0.818 | **0.859** |
| SR-ARE | 0.84 | NA | **0.850** |
| SR-HSE | 0.79 | NA | **0.892** |
| SR-MMP | 0.90 | NA | **0.943** |
| SR-P53 | 0.84 | NA | **0.910** |
| SR-ATAD5 | **0.84** | NA | 0.832 |

Table 7. Regression Co-efficient Performance ($R^2$) Comparison of Our Method with AdmetSAR on Cross-Validation for IGC50, LC50, LD50, and LD50DM

| task | AdmetSAR | our method |
|---|---|---|
| IGC50 | 0.822 | **0.825** |
| LC50-DM | NA | **0.616** |
| LC50 | 0.574 | **0.678** |
| LD50 | 0.613 | **0.629** |

**Feature Interpretability.** Machine learning models predominantly behave as "black box" which usually does not provide any explanation of the decisions made. In this study, we tried to interpret the outcome in terms of feature importance. For this, physicochemical 2D descriptors calculated using PaDEL package were used to build the predictive model. These features were ranked based on their gini index in the DT classifier. Gini index for individual toxicity task (7 NR tasks and 5 SR tasks) was calculated and added up to get the cumulative gini index to assign a single score to each feature across NR and SR toxicity data sets.

Figure 7a shows the cumulative gini index of 1422 features for NR and SR data sets. These features are arranged in a descending order of their gini index, the top 29 features in this list showed vertical drops in their gini index values, thus suggesting substantial difference in their importance, while others showed small variances (shown as the break point in Figure 7a). Similarly, average rank of each feature was calculated across NR and SR data sets. The relation between the average rank and the cumulative gini index score is shown in Figure 7b. The proportional behavior between these parameters confirms a consistent nature of features as per their importance score among all toxicity tasks of NR and SR data sets. Later, the top 29 gini index descriptors detected in gini index plot were identified separately. These 29 features with their average ranks are shown in Figure 7c. Here, it is observed that "path count descriptor" class is the most abundant class in the top features list (see Table S1 for the descriptors class detail). The top 3 features showed average rank below 10 are (1) pipC10, (2) pipC9, and (3) pipC8, and their average ranks are 3.91, 9.91, and 6.41, respectively (marked with red stars in Figure 7c). These 3 features from the path count descriptor class played the most critical role in classifying the molecule as toxic and nontoxic for NR and SR data sets. Relatively higher importance of these descriptors made them appropriate for coarse initial screening of molecules. In order to observe the importance of these top features, piPC10 values are plotted against all molecules. Figure 8 shows piPC10 values for toxic and nontoxic molecules, light blue circles represent toxic molecules while dark blue represent nontoxic molecules. As it can be clearly observed in Figure 8, toxic molecules make cluster in a certain range of piPC10 value, leaving a large area as safe zone (nontoxic). This shows the classifying property of piPC10 between the toxic and nontoxic molecule around a fixed value. The DT classifier used for feature importance in the presented hybrid framework has assigned a cut-off value to each feature at every node of the tree. These cut-offs of top features could be used as discriminating planes for toxic molecules.

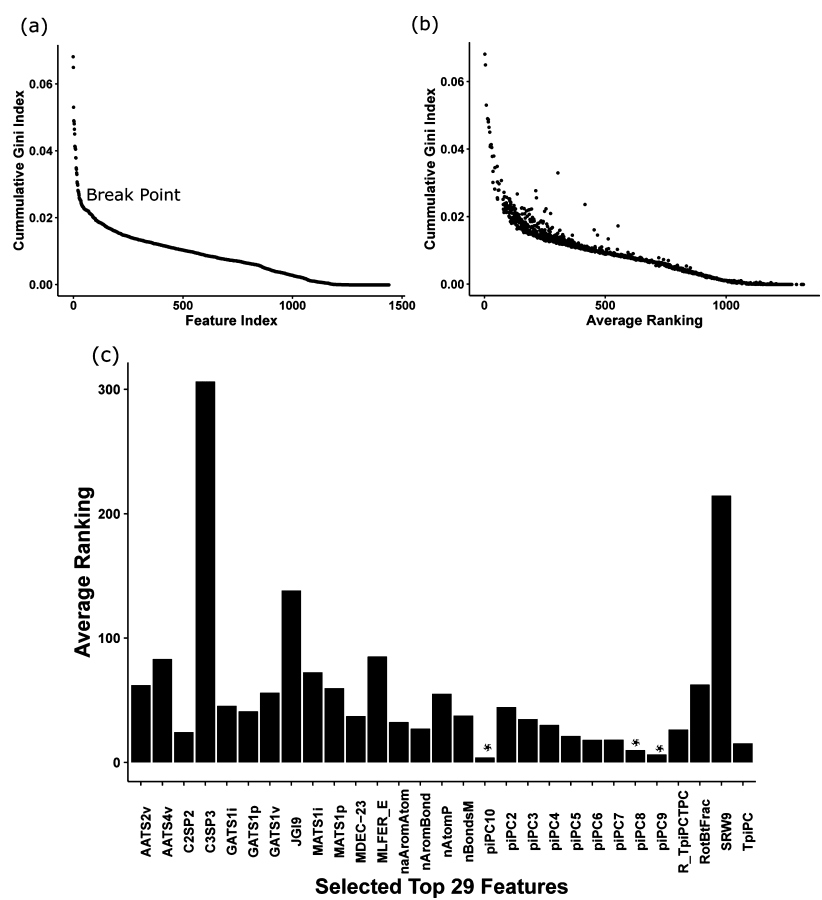

Figure 7. (a) Cumulative gini index score of 1422 features across 12 NR and SR toxicity data sets; (b) average ranking of 1422 features against cumulative gini index score in all 12 NR and SR data sets; and (c) ranking of top 29 features arranged in alphabetical order, top 3 features piPC10, piPC9, and piPC8 showed average rank below 10 and are marked with red star.

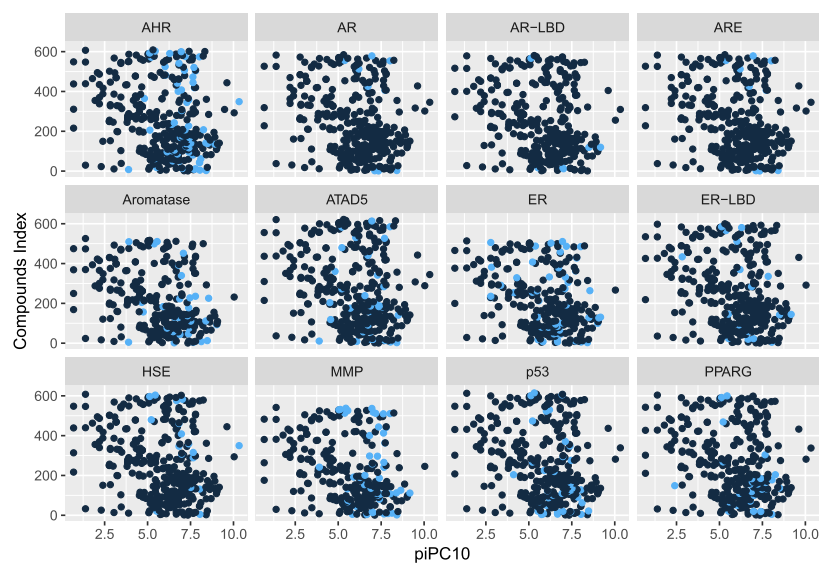

Figure 8. Classification of toxic and nontoxic molecule based on cut-off values of pipC10 features derived from DT classifier. Toxic molecules are shown in light blue while nontoxic are represented dark blue dots.

Feature cut-offs in DTs are defined as the values that divides the population in the highest ratios. Each tree has its own cut-off for each feature. Average cut-off values across these 1000 trees grown in building model for 3 most important features were calculated. These average cut-off values are given in Table S3. It is suggested that any molecule has a value for these descriptors less than the respective cut-off would have more possibility to be found in the toxic spectrum. It is shown in Table S3 that pipC10, pipC9, and pipC8 have similar cut-off ranges as they belong to the same descriptor class and showed similar behaviors in classifying the molecules. Top 3 features with their respective cut-offs were combined together to improve the discriminating power. Molecules that have values of these top 3 features less than their respective cut-offs are taken in one group. Table 8 shows that this group has less than

**Table 8. Toxic and Nontoxic Molecule Fraction Using Combined Criteria of pipC10, pipC9, and pipC8 for NR, SR, and AM**

| task | toxic molecule fraction | nontoxic molecule fraction |
| --- | --- | --- |
| NR-AHR | 0.01 | 0.48 |
| NR-AR | 0.01 | 0.47 |
| NR-ARLBD | 0.00 | 0.55 |
| NR-aromatase | 0.02 | 0.55 |
| NR-ER | 0.02 | 0.43 |
| NR-ERLBD | 0.01 | 0.44 |
| NR-PPARG | 0.01 | 0.48 |
| SR-ARE | 0.00 | 0.44 |
| SR-HSE | 0.01 | 0.45 |
| SR-MMP | 0.02 | 0.51 |
| SR-P53 | 0.00 | 0.49 |
| SR-ATAD5 | 0.01 | 0.51 |
| AM | 0.05 | 0.10 |

0.03 fraction toxic molecules to the total available toxic molecules for all 12 tasks of SR and NR while on an average 0.50 fraction nontoxic molecule to the total nontoxic molecules for respective classes. This suggests that combined criteria for piPC10, piPC9, and piPC8 could be used to find the probability of a given molecule to be toxic or nontoxic for SR and NR. Similarly, individual cut-offs of pipC10, pipC9, and pipC8 are 5.29, 5.10, and 5.0 for AM dataset. Later, these features and their respective cutoffs were used cumulatively on

AM data set as we done for NR/SR dataset. Here, again toxic molecules have low fraction 0.05 below the combined cut-off, whereas nontoxic molecules have 0.10 fraction. Although the fractional discrimination between toxic and nontoxic molecules on AM data is weaker that NR and SR dataset, but it clearly shows that piPC10, piPC9, and piPC8 can be used to determine the initial AM probability of any new molecules. Thus, these features could be used as initial indicators during molecule assessment.

## ■ DISCUSSION

Simplicity-accuracy trade-off of models for cheminformatics tasks is one of the most important factor to be considered before deploying a model. The proposed hybrid model is a step toward model simplicity and less compute intensiveness while still maintaining similar or higher accuracy to the DNN models for moderate size toxicity data sets. Here, we want to mention that after experimenting with various toxicity data sets of different size, and we found that our model struggles to achieve better accuracy for small data sets such as LC50 and LC50-DM as compared to the consensus and multi task models, though it comparable results to ST models. Generally, MT takes many tasks together and consensus takes many different models together to achieve higher performance, but on the other hand the complexity becomes very high. Predictive support is taken from other tasks or other models which hides the robustness and power of an individual model. In our case, we always take standalone ST model to compare against all the available methods. For smaller data sets, the feature selection method in our model selects relatively higher number of features which have high correlations among them. This is happening because the feature selection method in our model is acting as a coarse granular filter for small data sets. Currently, the better optimization of feature selection module is not in the scope of this paper and can be included in the future work.

Mostly, a large number of various types of features are computed and then used to predict toxicity end points. The range of these features is usually several thousands and that makes the model very complex and compute-intensive. These various types of features can be very hard to compute and that creates another bottleneck in toxicity end points prediction. We brought a two-level complexity reduction and tested our hybrid model on various toxicity data sets. The first level is related the simple and easy calculation of features. We used only 2D features which are relatively easier to calculate than 3D and other types of physicochemical features. The second level is using a simple SNN with only 1 hidden layer of 10 neurons, which is optimized jointly with DTs. This enabled us to further reduce the features and effectively select the subset of features. Here, it should be noted that our hybrid model selects the features automatically without human intervention and decides on effective number of features which produces better results, though the criteria of selecting the features can be further improved by using different heuristics. We did three case studies to show that for various toxicity end points, simple models would equally produce better results as compared to the complex models if fed with effective 2D features. We see our work in alignment with the three very important theories such as curse of dimensionality,[39] occam razor,[74] and universal approximation theorem.[63]

The simple model along with the cumulative feature ranking applied on classification toxicity tasks opens a venue of getting insights of the model predictions, though there is quite a room that the interpretation can be improved further. For instance, a DNN creates its own features by applying nonlinear function at every level to the input features, which are not human understandable specifically in chemistry or biology. In our case, as we effectively fed the SNN through an automatic feature selection, that makes it relatively easier to extract the decision rules from it.

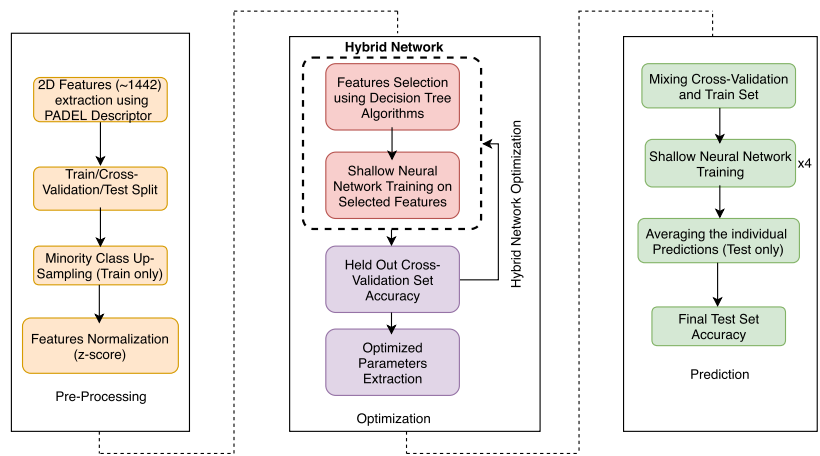

**Figure 9.** Prediction model flowchart.

## ■ CONCLUSIONS

Toxicity prediction of chemical compounds lately achieved significant progress in accuracy. The key factors behind this include using a huge set of features, implementing a complex blackbox technique such as a DNN, and exploiting enormous computational resources. In this paper, we strongly argue for the models and methods that are simple in machine learning characteristics, efficient in computing resource usage, and powerful to achieve very high accuracy levels. We developed and demonstrated a novel hybrid framework based on the joint optimization of DT and a SNN. Using this hybrid framework, we then build prediction model for three classification and four regression toxicity data sets. The joint optimization of DT and a SNN enabled us to achieve better or nearly similar results for various classification- and regression-based toxicity data sets. The model complexity as well as the training time is reduced by a large extent. Instead of utilizing thousands of features, only selected reduced number of important features made the model more comprehensible. This hybrid method reduces the dimensionality curse by using only reduced effective features. One of the aims of this study is to achieve comparable toxicity prediction results by using simpler machine learning model. This opens an avenue to highlight the insight of a prediction process in order to understand the specific problem in a comprehensive manner.

In our hybrid framework, a coarse filter for feature selection in the form of a DT prior to a prediction model based on gini-index was applied. DTs helped in feature analysis using cumulative gini index. This was performed to find global relevance of features across toxicity tasks of SR and NR. Additionally, individual rankings of these features were used to calculate average ranking of each feature. The correlation between the average rank and cumulative gini index suggests the similar importance pattern of these features among diverse toxicity tasks. Eventually, the top features based on the gini index were plotted and 3 features were observed. (1) pipC10, (2) pipC9, and (3) pipC8 have average ranks below 10. They belong to single descriptor class called path count. There individual cut-offs at first node were extracted from 1000 DTs and average score was used to observe the classification potential of these top features on toxic and nontoxic compounds. piPC10 was initially plotted for all toxicity tasks and clear discrimination was observed between toxic and nontoxic molecules for SR and NR. Further, piPC9 and piPC8 were combined with piPC10 to design cumulative criteria for classification. The cumulative criteria indicate a safe zone, where the probability of finding toxic compounds is less than 0.05%. This can allow users for initial screening of toxic and nontoxic compounds based on only piPC10, piPC9, and piPC8 scores.

We conclude that our hybrid model of a DT and an SNN can be used for toxicity prediction or any similar tasks to achieve better or near similar accuracy in comparably lesser time and lesser resources. This technique enabled us to use certain features for rapid and prior toxicity estimation. It will also be interesting to apply a coarse feature selection method using a heuristic approach to improve feature space optimization. Following are the main concluding points of our study.

- We propose an efficient hybrid algorithm which effectively selects a feature subset of 2D features for training.
- The use of significantly reduced number of effective 2D features helps in interpretability.
- The computational complexity of various toxicity end points can be reduced to a great extent with our hybrid algorithm while keeping the accuracy level similar or relatively better than the state of the art methods.
- Using our commutative feature ranking method, we help the chemists effectively screen out the toxic compounds with few features in hand.

## ■ METHODS

The work flow of our hybrid framework is composed of three main blocks as shown in Figure 9. All these three main blocks with submodules in each are explained below.

**Preprocessing.** In preprocessing, 1422 2D chemical descriptors for NR and SR while 1249 for AM were calculated using an open source package called PADEL (a list is given in Table S1).[60] Data are split into train, CV, and test sets. The split for NR and SR was predefined by the Tox21 challenge,[75] where a separate held out CV set of ≈296 instances is provided for an in-house CV purpose. On AM data, no such division was given so we divided it into 60% train, 20% CV, and 20% test sets. The train and test sets for NR and SR consists of ≈8000 and ≈647 unique instances, respectively. In order to avoid biasness of the model toward majority instances, minority class was up-sampled. Data were also normalized using a data scaling method. We provide the important pieces of python code for reproducibility.

1 Extracting 2D features

PaDEL-Descriptor.jar -removesalt -standardizenitro -standardizetautomers -2d

2 Data split (If the data split is not given already)

X_train_cv, X_test, y_train_cv, y_test = train_test_split­(X,y,test_size = 0.20)#20% test set X_train, X_cv, y_train, y_cv = train_test_split(X_train_cv,y_train_cv,test_size = 0.20)#20% cv set

3 Minority class up-sampling (if required):

train_minority_upsampled = resample(train_minority, replace = True, n_samples = train_majority_size) train_upsampled = pd.concat([train_majority, train_minority_upsampled])

4 Z-Score normalization

Data_input_norm = (Data_input - Data_input_mean)/Data_input_std

**Hybrid Framework.** Considering a feature selection approach, we designed a novel hybrid framework that consists of two components: a DT and a SNN. DTs acted as a coarse filter to select a reduced number of features in order to train the SNN. DTs with feature selection technique helps in interpretability and provides with a criterion for prescreening the compounds in all three toxicity data sets while SNN helps improve the accuracy. Training with selected feature subspace reduces time and model complexity which leads to better interpretability.[76]

**Optimization.** In model optimization, both components (DT and the SNN) of hybrid framework were conjointly optimized. Here the chosen objective function of a neural network is dependent on its own parameters as well as on parameters of the feature selection module. A held out predefined CV set was used to optimize both components of the hybrid model as discussed below:

*Feature Selection via DT.* In feature selection module, we used an extremely randomized ET classifier (a type of DT)[61] with gini index, also called mean decrease impurity[77,78] to perform initial coarse filtering for features ranking.[79] As our aim is to tweak the number of selected features, so only those parameters were optimized that affect the process of selecting the features. The ET classifier has several optimization parameters but the most critical ones are (1) n_estimators that represent the number of trees in the forest and (2) threshold that limits the number of features selected during optimization.[62] All the features were ranked on the basis of the gini index. The higher the gini index value, the greater the importance of that feature in predicting a specific class.[79] We provide the important pieces of python code for reproducibility.

#calling the hybrid model for optimization
hybrid_model = hybrid_model_opt()
#hybrid_model_opt() is given in Supporting Information

#Optimizing the threshold value while keeping SNN parameters fixed param_grid = {'fs__threshold': ['0.08*mean','0.09*mean','0.10*mean', '0.2*mean','0.3*mean','0.4*mean','0.5*mean','0.6*mean','0.7*mean', '0.8*mean','0.9*mean','1*mean','1.1*mean','1.2*mean','1.3*mean', '1.4*mean','1.5*mean','1.6*mean','1.7*mean','1.8*mean','1.9*mean', '2.0*mean','2.1*mean','2.2*mean','2.3*mean'], 'clf__dropout_rate': [0.5],'clf__epochs': [20], 'clf__batch_size': [512],'clf__init_mode': ['he-normal'], 'clf__activation': ['relu']}

#Grid search
grid = GridSearchCV(estimator = hybrid_model, param_grid = param_grid, scoring = 'accuracy',cv = 5)
opt_result = grid.fit(train_cv_x, train_cv_y)

During feature selection process via threshold parameter optimization, parameters of the SNN were fixed as shown in Table 9. Because of the single parameter optimization, a grid search was applied on threshold value to achieve maximum accuracy. A higher value of the threshold reflects a smaller number of features, whereas a lower value reflects a large number of available features. The range of the threshold for grid search was set such that it can select a small number of features up to the all available features.

Table 9. Hybrid Model Feature Selection Optimization

| threshold (grid search) | | | | |
|---|---|---|---|---|
| 0.08 × mean | 0.09 | 0.1 | 0.2 | 0.3 |
| 0.4 | 0.5 | 0.6 | 0.7 | 0.8 |
| 0.9 | 1 | 1.1 | 1.2 | 1.3 |
| 1.4 | 1.5 | 1.6 | 1.7 | 1.8 |
| 1.9 | 2 | 2.1 | 2.2 | 2.3 |
| fixed parameters | | | | |
| epochs | 20 | | | |
| initialization function | He-normal | | | |
| DropOut | 0.5 | | | |
| activation | Relu | | | |
| mini-batch | 512 | | | |

*SNN Hyper-Parameters Tuning.* Once the reduced feature subspace was obtained in the feature selection process, then with the selected features, hyper-parameters were tuned for the SNN as shown in Table 10. Then, a random search was performed for SNN hyper-parameters tuning because it is more efficient than the grid search in case of more parameters to optimize.[80] We provide the important pieces of python code for reproducibility.

Table 10. Hybrid Model SNN Optimization[a]

| epochs | 10, 20, 40, 50, 60, 200, 250, 400 |
|---|---|
| initialization function | He-normal, He-uniform normal, uniform, Glorot-normal |
| DropOut | 0.0, 0.1, 0.2, 0.3, 0.4, 0.5, 0.6, 0.7, 0.8, 0.9 |
| activation | Relu, sigmoid |
| mini-batch | 32, 64, 128, 512, 1024, 2048, 4096, 8192 |

[a]SNN hyper-parameter tuning (random search).

#calling the hybrid model for optimization
hybrid_model = hybrid_model_opt()
#hybrid_model_opt() is given in Supporting Information S5.

#Optimizing the SNN parameters

# fs instance is used for feature selection module of hybrid model
# clf is used for SNN module
param_grid = {'fs__threshold':['opt_result.best_params_["fs__threshold"]'],
 'clf__dropout_rate': [0.1, 0.2, 0.3, 0.4, 0.5, 0.6, 0.7, 0.8, 0.9],
 'clf__epochs': [10, 20, 40, 50, 60, 200, 250, 400],
 'clf__init_mode': ['uniform','lecun_uniform','normal','glorot_normal', 'he_normal','he_uniform'], 'clf__batch_size': [32, 64, 128, 512, 1024, 2048, 4096, 8192], 'clf__activation': ['relu','sigmoid']}
#Random search
grid = RandomizedSearchCV((estimator = hybrid_model, param_distributions = param_grid, n_iter = 50,scoring = 'accuracy',cv = 5) opt_result = grid.fit(train_cv_x, train_cv_y)

**Training and Prediction.** In prediction, the CV and the training set were mixed together after obtaining all the optimized parameters. Optimized parameters were used to train the SNN for each individual toxicity task of all data sets. A set of four similar SNNs were trained and their outputs were averaged to form a more robust model. The detail of the optimized parameters for all three toxicity data sets and the result of each ensemble network are given in Table S2. Complete pipeline of hybrid prediction framework is shown in Figure 9. We provide the important pieces of python code for reproducibility.

# Calling the model_nn_final with optimized parameters
# model_nn_final() is given in Supporting Information S5
#selected_x and selected_y are the data sets containing-
#-only selected features
pred_test = model_nn_final(train_selected_x, train_selected_y,
 test_selected_x, test_selected_y,
 opt_result.best_params_["clf__dropout_rate"], opt_result.best_params_["clf__epochs"], opt_result.best_params_["clf__init_mode"], opt_result.best_params_["clf__batch_size"], opt_result.best_params_["clf__activation"]).

## ■ ASSOCIATED CONTENT

**ⓢ Supporting Information**
The Supporting Information is available free of charge on the ACS Publications website at DOI: 10.1021/acsomega.8b03173.

> All the 2D features and their class type for all toxicity data sets considered in this study; optimized parameters for all the tasks; cut-off value list for top 3 features pipC10, pipC9, and pipc8 from path count descriptor class for NR and SR toxicity tasks; the exact 2D features which were selected for each task; codes for hybrid_model_opt() and model_nn_final() (data and code availability: the trained models along with the data sets for SR, NR, AM, IGC50, LD50, LC50-DM, and LC50 are available at HybridTox2D. We also provide the source code to reproduce the results of this paper. We provide the original links to the data sets as follows. SR, NR data sets: Tox21 Data Challenge 2014. AM benchmark data set: toxbenchmark. IGC50, LD50, LC50-DM, and LC-50 data sets: these data sets can be obtained from the authors of TopTox) (PDF)

## ■ AUTHOR INFORMATION

**Corresponding Authors**
*E-mail: abdul.karim@griffithuni.edu.au (A.K.).
*E-mail: avish2k@gmail.com (A.M.).
*E-mail: mahakim.newton@griffith.edu.au (M.A.H.N.).
*E-mail: a.sattar@griffith.edu.au (A.S.).

**ORCID**
Abdul Karim: 0000-0002-4431-7507
Avinash Mishra: 0000-0003-4125-6670

**Author Contributions**
§A.K. and A.M. contributed equally.

**Notes**
The authors declare no competing financial interest.

## ■ ACKNOWLEDGMENTS

The authors acknowledge support from the Institute for Integrated and Intelligent Systems, Griffith University and the Department of Biotechnology (DBT), India for the award of an Indo-Australian Gold fellowship. The authors acknowledge Kedi Wu and Guo-Wei from Michigan State University for providing additional toxicity data sets for regression models.

## ■ REFERENCES

(1) Sorell, T. L. Approaches to the development of human health toxicity values for active pharmaceutical ingredients in the environment. *AAPS J.* **2015**, *18*, 92−101.
(2) Olson, H.; Betton, G.; Robinson, D.; Thomas, K.; Monro, A.; Kolaja, G.; Lilly, P.; Sanders, J.; Sipes, G.; Bracken, W.; et al. Concordance of the toxicity of pharmaceuticals in humans and in animals. *Regul. Toxicol. Pharmacol.* **2000**, *32*, 56−67.
(3) Zhu, H.; Tropsha, A.; Fourches, D.; Varnek, A.; Papa, E.; Gramatica, P.; Öberg, T.; Dao, P.; Cherkasov, A.; Tetko, I. V. Combinatorial QSAR modeling of chemical toxicants tested against Tetrahymena pyriformis. *J. Chem. Inf. Model.* **2008**, *48*, 766−784.
(4) Vedani, A.; Dobler, M.; Lill, M. A. The challenge of predicting drug toxicity in silico. *Basic Clin. Pharmacol. Toxicol.* **2006**, *99*, 195−208.
(5) Dekant, W.; Melching-Kollmuß, S.; Kalberlah, F. Toxicity assessment strategies, data requirements, and risk assessment approaches to derive health based guidance values for non-relevant metabolites of plant protection products. *Regul. Toxicol. Pharmacol.* **2010**, *56*, 135−142.
(6) Parasuraman, S. Toxicological screening. *J. Pharmacol. Pharmacother.* **2011**, *2*, 74−79.
(7) Curren, R. D.; Harbell, J. W. In vitro alternatives for ocular irritation. *Environ. Health Perspect.* **1998**, *106*, 485.
(8) York, M.; Steiling, W. A critical review of the assessment of eye irritation potential using the Draize rabbit eye test. *J. Appl. Toxicol.* **1998**, *18*, 233−240.
(9) Eastmond, D. A.; Hartwig, A.; Anderson, D.; Anwar, W. A.; Cimino, M. C.; Dobrev, I.; Douglas, G. R.; Nohmi, T.; Phillips, D. H.; Vickers, C. Mutagenicity testing for chemical risk assessment: update of the WHO/IPCS Harmonized Scheme. *Mutagenesis* **2009**, *24*, 341−349.
(10) Gholami, S.; Soleimani, F.; Shirazi, F. H.; Touhidpour, M.; Mahmoudian, M. Evaluation of mutagenicity of mebudipine, a new calcium channel blocker. *Iranian Journal of Pharmaceutical Research: IJPR* **2010**, *9*, 49.
(11) Zepnik, H.; Völkel, W.; Dekant, W. Toxicokinetics of the mycotoxin ochratoxin A in F 344 rats after oral administration. *Toxicol. Appl. Pharmacol.* **2003**, *192*, 36−44.
(12) Ema, M.; Fukui, Y.; Aoyama, H.; Fujiwara, M.; Fuji, J.; Inouye, M.; Iwase, T.; Kihara, T.; Oi, A.; Otani, H.; et al. Comments from the developmental neurotoxicology committee of the Japanese teratology society on the OECD guideline for the testing of chemicals, proposal for a new guideline 426, developmental neurotoxicity study, draft document (October 2006 version), and on the draft document of the retrospective performance assessment of the draft test guideline 426 on developmental neurotoxicity. *Congenital Anomalies* **2007**, *47*, 74−76.

(13) Kimm-Brinson, K. L.; Ramsdell, J. S. The red tide toxin, brevetoxin, induces embryo toxicity and developmental abnormalities. *Environ. Health Perspect.* **2001**, *109*, 377.

(14) Oliveira, C. D. R.; Moreira, C. Q.; de Sá, L. R. M.; Spinosa, H. d. S.; Yonamine, M. Maternal and developmental toxicity of ayahuasca in Wistar rats. *Birth Defects Res., Part B* **2010**, *89*, 207−212.

(15) Strovel, J.; Sittampalam, S.; Coussens, N. P.; Hughes, M.; Inglese, J.; Kurtz, A.; Andalibi, A.; Patton, L.; Austin, C.; Baltezor, M.; et al. *Early Drug Discovery and Development Guidelines: For Academic Researchers, Collaborators, and Start-Up Companies*, National Center for Advancing Translational Sciences (NCATS): 2016.

(16) Rowan, A. N. Ending the Use of Animals in Toxicity Testing and Risk Evaluation. *Camb. Q. Healthc. Ethics* **2015**, *24*, 448−458.

(17) Andersen, M. E.; Krewski, D. Toxicity testing in the 21st century: bringing the vision to life. *Toxicol. Sci.* **2008**, *107*, 324−330.

(18) Rusyn, I.; Daston, G. P. Computational toxicology: realizing the promise of the toxicity testing in the 21st century. *Environ. Health Perspect.* **2010**, *118*, 1047.

(19) Xia, M.; Huang, R.; Witt, K. L.; Southall, N.; Fostel, J.; Cho, M.-H.; Jadhav, A.; Smith, C. S.; Inglese, J.; Portier, C. J.; et al. Compound cytotoxicity profiling using quantitative high-throughput screening. *Environ. Health Perspect.* **2008**, *116*, 284.

(20) Lima, A. N.; Philot, E. A.; Trossini, G. H. G.; Scott, L. P. B.; Maltarollo, V. G.; Honorio, K. M. Use of machine learning approaches for novel drug discovery. *Expert Opin. Drug Discovery* **2016**, *11*, 225−239.

(21) Chavan, S.; Friedman, R.; Nicholls, I. Acute toxicity-supported chronic toxicity prediction: a k-nearest neighbor coupled read-across strategy. *Int. J. Mol. Sci.* **2015**, *16*, 11659−11677.

(22) Kauffman, G. W.; Jurs, P. C. QSAR and k-nearest neighbor classification analysis of selective cyclooxygenase-2 inhibitors using topologically-based numerical descriptors. *J. Chem. Inf. Comput. Sci.* **2001**, *41*, 1553−1560.

(23) Ajmani, S.; Jadhav, K.; Kulkarni, S. A. Three-dimensional QSAR using the k-nearest neighbor method and its interpretation. *J. Chem. Inf. Model.* **2006**, *46*, 24−31.

(24) Sakuratani, Y.; Zhang, H. Q.; Nishikawa, S.; Yamazaki, K.; Yamada, T.; Yamada, J.; Gerova, K.; Chankov, G.; Mekenyan, O.; Hayashi, M. Hazard Evaluation Support System (HESS) for predicting repeated dose toxicity using toxicological categories. *SAR QSAR Environ. Res.* **2013**, *24*, 351−363.

(25) Deng, C.-H.; Zhao, W.-L. Fast k-means based on KNN Graph. **2017**, arXiv preprint arXiv:1705.01813.

(26) Ailon, N.; Jaiswal, R.; Monteleoni, C. Streaming k-means approximation. *Advances in Neural Information Processing Systems*; MIT Press. 2009; pp 10−18.

(27) Vattani, A. K-means requires exponentially many iterations even in the plane. *Discrete Comput. Geom.* **2011**, *45*, 596−616.

(28) Svetnik, V.; Liaw, A.; Tong, C.; Culberson, J. C.; Sheridan, R. P.; Feuston, B. P. Random forest: a classification and regression tool for compound classification and QSAR modeling. *J. Chem. Inf. Comput. Sci.* **2003**, *43*, 1947−1958.

(29) Xia, X.; Maliski, E. G.; Gallant, P.; Rogers, D. Classification of kinase inhibitors using a Bayesian model. *J. Med. Chem.* **2004**, *47*, 4463−4470.

(30) Bender, A.; Mussa, H. Y.; Glen, R. C.; Reiling, S. Molecular similarity searching using atom environments, information-based feature selection, and a naive Bayesian classifier. *J. Chem. Inf. Comput. Sci.* **2004**, *44*, 170−178.

(31) Polishchuk, P. G.; Muratov, E. N.; Artemenko, A. G.; Kolumbin, O. G.; Muratov, N. N.; Kuz'min, V. E. Application of random forest approach to QSAR prediction of aquatic toxicity. *J. Chem. Inf. Model.* **2009**, *49*, 2481−2488.

(32) Tong, W.; Hong, H.; Fang, H.; Xie, Q.; Perkins, R. Decision forest: combining the predictions of multiple independent decision tree models. *J. Chem. Inf. Comput. Sci.* **2003**, *43*, 525−531.

(33) Unterthiner, T.; Mayr, A.; Klambauer, G.; Hochreiter, S. Toxicity prediction using deep learning. **2015**, arXiv preprint arXiv:1503.01445.

(34) Bengio, Y.; et al. Learning deep architectures for AI. *Found. Trends Mach. Learn.* **2009**, *2*, 1−127.

(35) Schmidhuber, J. Deep learning in neural networks: An overview. *Neural Networks* **2015**, *61*, 85−117.

(36) Mhaskar, H.; Liao, Q.; Poggio, T. A. When and why are deep networks better than shallow ones? *AAAI, Thirty-First AAAI Conference on Artificial Intelligence* **2017**, 2343−2349.

(37) Winkler, D. A.; Le, T. C. Performance of deep and shallow neural networks, the universal approximation theorem, activity cliffs, and QSAR. *Mol. Inf.* **2017**, *36*, 1600118.

(38) Nasrabadi, N. M. Pattern recognition and machine learning. *J. Electron. Imag.* **2007**, *16*, 049901.

(39) Trunk, G. V. A problem of dimensionality: A simple example. *IEEE Trans. Pattern Anal. Mach. Intell.* **1979**, *PAMI-1*, 306−307.

(40) Dauphin, Y. N.; Bengio, Y. Big neural networks waste capacity. **2013**, arXiv preprint arXiv:1301.3583.

(41) Ma, J.; Sheridan, R. P.; Liaw, A.; Dahl, G. E.; Svetnik, V. Deep neural nets as a method for quantitative structure−activity relationships. *J. Chem. Inf. Model.* **2015**, *55*, 263−274.

(42) Kato, Y.; Hamada, S.; Goto, H. Molecular activity prediction using deep learning software library. *Advanced Informatics: Concepts, Theory And Application (ICAICTA), 2016 International Conference On*, 2016; pp 1−6.

(43) Xu, Y.; Dai, Z.; Chen, F.; Gao, S.; Pei, J.; Lai, L. Deep learning for drug-induced liver injury. *J. Chem. Inf. Model.* **2015**, *55*, 2085−2093.

(44) Montavon, G.; Rupp, M.; Gobre, V.; Vazquez-Mayagoitia, A.; Hansen, K.; Tkatchenko, A.; Müller, K.-R.; von Lilienfeld, O. A. Machine learning of molecular electronic properties in chemical compound space. *New J. Phys.* **2013**, *15*, 095003.

(45) Dahl, G. E.; Jaitly, N.; Salakhutdinov, R. Multi-task neural networks for QSAR predictions. **2014**, arXiv preprint arXiv:1406.1231.

(46) Ma, J.; Sheridan, R. P.; Liaw, A.; Dahl, G. E.; Svetnik, V. Deep neural nets as a method for quantitative structure−activity relationships. *J. Chem. Inf. Model.* **2015**, *55*, 263−274.

(47) Ramsundar, B.; Kearnes, S.; Riley, P.; Webster, D.; Konerding, D.; Pande, V. Massively multitask networks for drug discovery. **2015**, arXiv preprint arXiv:1502.02072.

(48) Hughes, T. B.; Le Dang, N.; Miller, G. P.; Swamidass, S. J. Modeling reactivity to biological macromolecules with a deep multitask network. *ACS Cent. Sci.* **2016**, *2*, 529−537.

(49) Hughes, T. B.; Miller, G. P.; Swamidass, S. J. Modeling epoxidation of drug-like molecules with a deep machine learning network. *ACS Cent. Sci.* **2015**, *1*, 168−180.

(50) Hughes, T. B.; Miller, G. P.; Swamidass, S. J. Site of reactivity models predict molecular reactivity of diverse chemicals with glutathione. *Chem. Res. Toxicol.* **2015**, *28*, 797−809.

(51) Lusci, A.; Pollastri, G.; Baldi, P. Deep architectures and deep learning in chemoinformatics: the prediction of aqueous solubility for drug-like molecules. *J. Chem. Inf. Model.* **2013**, *53*, 1563−1575.

(52) Kearnes, S.; Goldman, B.; Pande, V. Modeling industrial admet data with multitask networks. **2016**, arXiv preprint arXiv:1606.08793.

(53) Wallach, I.; Dzamba, M.; Heifets, A. Atomnet: A deep convolutional neural network for bioactivity prediction in structure-based drug discovery. **2015**, arXiv preprint arXiv:1510.02855.

(54) Smith, J. S.; Isayev, O.; Roitberg, A. E. ANI-1: an extensible neural network potential with DFT accuracy at force field computational cost. *Chem. Sci.* **2017**, *8*, 3192−3203.

(55) Schütt, K. T.; Arbabzadah, F.; Chmiela, S.; Müller, K. R.; Tkatchenko, A. Quantum-chemical insights from deep tensor neural networks. *Nat. Commun.* **2017**, *8*, 13890.

(56) Svetnik, V.; Liaw, A.; Tong, C.; Culberson, J. C.; Sheridan, R. P.; Feuston, B. P. Random forest: a classification and regression tool for compound classification and QSAR modeling. *J. Chem. Inf. Comput. Sci.* **2003**, *43*, 1947−1958.

(57) Mayr, A.; Klambauer, G.; Unterthiner, T.; Hochreiter, S. DeepTox: toxicity prediction using deep learning. *Front. Environ. Sci.* **2016**, *3*, 80.

(58) Huang, R.; Sakamuru, S.; Martin, M. T.; Reif, D. M.; Judson, R. S.; Houck, K. A.; Casey, W.; Hsieh, J.-H.; Shockley, K. R.; Ceger, P.; et al. Profiling of the 10K compound library for agonists and antagonists of the estrogen receptor alpha signaling pathway. *Sci. Rep.* **2014**, *4*, 5664.

(59) Hansen, K.; Mika, S.; Schroeter, T.; Sutter, A.; ter Laak, A.; Steger-Hartmann, T.; Heinrich, N.; Müller, K.-R. Benchmark data set for in silico prediction of Ames mutagenicity. *J. Chem. Inf. Model.* **2009**, *49*, 2077−2081.

(60) Yap, C. W. PaDEL-descriptor: An open source software to calculate molecular descriptors and fingerprints. *J. Comput. Chem.* **2010**, *32*, 1466−1474.

(61) Geurts, P.; Ernst, D.; Wehenkel, L. Extremely randomized trees. *Mach. Learn.* **2006**, *63*, 3−42.

(62) Pedregosa, F.; Varoquaux, G.; Gramfort, A.; Michel, V.; Thirion, B.; Grisel, O.; Blondel, M.; Prettenhofer, P.; Weiss, R.; Dubourg, V.; et al. Scikit-learn: Machine learning in Python. *Journal of Machine Learning Research* **2011**, *12*, 2825−2830.

(63) Csáji, B. C. Approximation with artificial neural networks. MSc Thesis, Faculty of Sciences, Etvs Lornd University, Hungary, 2001, 24, 48.

(64) Abdelaziz, A.; Spahn-Langguth, H.; Schramm, K.-W.; Tetko, I. V. Consensus modeling for HTS assays using in silico descriptors calculates the best balanced accuracy in Tox21 challenge. *Front. Environ. Sci.* **2016**, *4*, 2.

(65) Barta, G. Identifying Biological Pathway Interrupting Toxins Using Multi-Tree Ensembles. *Front. Environ. Sci.* **2016**, *4*, 52.

(66) Capuzzi, S. J.; Politi, R.; Isayev, O.; Farag, S.; Tropsha, A. QSAR modeling of Tox21 challenge stress response and nuclear receptor signaling toxicity assays. *Front. Environ. Sci.* **2016**, *4*, 3.

(67) Goh, G. B.; Hodas, N.; Siegel, C.; Vishnu, A. SMILES2vec: Predicting Chemical Properties from Text Representations, Workshop track - ICLR 2018: 2018.

(68) Goh, G. B.; Siegel, C.; Vishnu, A.; Hodas, N. O.; Baker, N. Chemception: A deep neural network with minimal chemistry knowledge matches the performance of expert-developed qsar/qspr models. **2017**, arXiv preprint arXiv:1706.06689.

(69) Banerjee, P.; Eckert, A. O.; Schrey, A. K.; Preissner, R. ProTox-II: a webserver for the prediction of toxicity of chemicals. *Nucleic Acids Res* **2018**, *46*, W257.

(70) Wu, K.; Wei, G.-W. Quantitative Toxicity Prediction Using Topology Based Multitask Deep Neural Networks. *J. Chem. Inf. Model.* **2018**, *58*, 520−531.

(71) Cheng, F.; Li, W.; Zhou, Y.; Shen, J.; Wu, Z.; Liu, G.; Lee, P. W.; Tang, Y. admetSAR: a comprehensive source and free tool for assessment of chemical ADMET properties. *J. Chem. Inf. Model.* **2012**, *52*, 3099.

(72) Yang, H.; Lou, C.; Sun, L.; Li, J.; Cai, Y.; Wang, Z.; Li, W.; Liu, G.; Tang, Y. admetSAR 2.0: web-service for prediction and optimization of chemical ADMET properties. *Bioinformatics* **2018**, bty707.

(73) Lea, I. A.; Gong, H.; Paleja, A.; Rashid, A.; Fostel, J. CEBS: a comprehensive annotated database of toxicological data. *Nucleic Acids Res.* **2016**, *45*, D964−D971.

(74) Blumer, A.; Ehrenfeucht, A.; Haussler, D.; Warmuth, M. K. Occam's Razor. *Inf. Process. Lett.* **1987**, *24*, 377−380.

(75) Huang, R.; Xia, M.; Nguyen, D.-T.; Zhao, T.; Sakamuru, S.; Zhao, J.; Shahane, S. A.; Rossoshek, A.; Simeonov, A. Tox21Challenge to build predictive models of nuclear receptor and stress response pathways as mediated by exposure to environmental chemicals and drugs. *Front. Environ. Sci.* **2016**, *3*, 85.

(76) Hill, M. O. *Introduction to the Exploration of Multivariate Biological Data*; Backhuys Publishers, 2002.

(77) Strobl, C.; Hothorn, T.; Zeileis, A. Party on! *The R Journal* **2009**, *1/2*, 14−17.

(78) Louppe, G.; Wehenkel, L.; Sutera, A.; Geurts, P. Understanding variable importances in forests of randomized trees. *Advances in Neural Information Processing Systems*; MIT Press, 2013; pp 431−439.

(79) Menze, B. H.; Kelm, B. M.; Masuch, R.; Himmelreich, U.; Bachert, P.; Petrich, W.; Hamprecht, F. A. A comparison of random forest and its Gini importance with standard chemometric methods for the feature selection and classification of spectral data. *BMC Bioinf.* **2009**, *10*, 213.

(80) Bergstra, J.; Bengio, Y. Random search for hyper-parameter optimization. *J. Mach. Learn. Res.* **2012**, *13*, 281−305.